# MACHINE LEARNING TECHNIQUES WITH ONTOLOGY FOR SUBJECTIVE ANSWER EVALUATION

M. Syamala Devi and Himani Mittal

Department of Computer Science and Applications, Panjab University, Chandigarh, India

## *ABSTRACT*

*Computerized Evaluation of English Essays is performed using Machine learning techniques like Latent Semantic Analysis (LSA), Generalized LSA, Bilingual Evaluation Understudy and Maximum Entropy. Ontology, a concept map of domain knowledge, can enhance the performance of these techniques. Use of Ontology makes the evaluation process holistic as presence of keywords, synonyms, the right word combination and coverage of concepts can be checked. In this paper, the above mentioned techniques are implemented both with and without Ontology and tested on common input data consisting of technical answers of Computer Science. Domain Ontology of Computer Graphics is designed and developed. The software used for implementation includes Java Programming Language and tools such as MATLAB, Protégé, etc. Ten questions from Computer Graphics with sixty answers for each question are used for testing. The results are analyzed and it is concluded that the results are more accurate with use of Ontology.*

## 1. INTRODUCTION

The manual system for evaluation of Subjective Answers for technical subjects involves a lot of time and effort of the evaluator. Performing evaluation through computers using intelligent techniques ensures uniformity in marking as the same inference mechanism is used for all the students. Subjective answers are evaluated on the basis of content and style of writing. For technical subjects, emphasis is more on content. If standard keywords are found in students' answer then answer is correct. However, we cannot mark the answers by just counting the number of keywords. A more wholesome approach is required, which can evaluate on the basis of not only keyword presence but the semantic relationship between words and concepts.

Several Machine Learning (ML) techniques exist that try to capture the latent relationship between words. The techniques explored in this paper are - Latent Semantic Analysis (LSA), Generalized Latent Semantic Analysis (GLSA), Maximum Entropy (MaxEnt) and BiLingual Evaluation Understudy (BLEU). These techniques are applied to Subjective Evaluation in previous work of other authors. The existing techniques show an accuracy of results from 59 to 93 percent. However, the evaluation is purely on the basis of keyword presence. An evaluation process is required which measures the wholesome relationship between words, words and concepts and among concepts. Ontology is a concept map of domain knowledge. In this paper, we explore the effect of using Ontology with ML techniques. All the above mentioned techniques are implemented both with and without Ontology and tested on common input data. The data was collected by conducting class tests of graduate and post-graduate classes. Then, Human evaluation of these answers was done. Correlations are calculated between human assigned marks and scores of each technique. The use of Ontology with these techniques ensures proper semantic based evaluation as the answers are matched against an exhaustive Knowledge Base. This ensures categorization of answers on the basis of concept category to which the answers belong. Ontology makes the evaluation process holistic as presence of keywords, synonyms, the right word combination and coverage of concepts can be checked. In addition to Subjective Evaluation, Ontology is also used in document categorization, plagiarism detection, sentiment analysis, information retrieval and information extraction.

The paper is organized as follows: Section 2 contains the review of related work. Section 3 and 4 state the algorithm steps followed to apply machine learning techniques to technical answer evaluation with and without Ontology. Section 5 discusses the implementation details of different techniques. In Section 6, testing and results are given. Section 7 includes analysis of results. Section 8 contains the conclusions of this research work.

## 2. Review Of Related Work

The research for evaluating subjective answers using computers is ongoing for more than a decade. Several Machine Learning techniques are applied to Subjective Answer Evaluation. Table-1 contains a survey of techniques and tools in Subjective evaluation, already used and future techniques that can be used. The techniques are classified in five categories- Clustering techniques, classification techniques, hybrid natural language processing techniques, soft computing techniques and Semantic techniques.

Latent Semantic Analysis (LSA) technique, proposed by Deerwester [23], is used to establish similarity between two contents. Intelligent Essay Assessor (IEA) tool uses LSA [3-7] for subjective evaluation of English essays. IEA was applied to TOEFL exam (ETS) and accuracy of results varies between 59 to 87 percent. Diana Perez [19-20] developed a tool-Atenea, using hybrid of LSA and BiLingual Evaluation Understudy (BLEU). The two techniques are applied independently and the results are combined by a linear equation. However, the coefficient of BLEU and LSA score in the linear equation is left as open question. Author has shown multiple combinations and average success rate is 50 percent of times. Electronic Essay Rater (E-Rater) [17] uses Natural Language Processing techniques to evaluate sentence structure. It is successfully used for AWA test in GMAT and has 84 to 93 percent accuracy. The set of features used are specific to AWA and TOEFL test; its applicability to technical answers is not proven. Discrepant essays are scored like regular essays. Generalized Latent Semantic Analysis (GLSA) [11] extends LSA. Instead of word by document matrix in LSA, phrase (n-gram groups of two or more words) by document matrix is created. Its success rate is 89 percent. C-rater [14-16] uses Maximum Entropy technique (MaxEnt). It has 80 percent agreement with the score assigned by a human-grader for short answers.

Most of the above techniques are information retrieval techniques of Natural Language Processing. However, the new research area in natural language processing is semantic mapping and information extraction. The Knowledge Base, Ontology is used for semantic mapping. Domain Ontology represents knowledge in the form of Subject-Object-Predicate triples. The application of Ontology to document classification is discussed in [1]. It uses bag-of-concepts approach instead of bag-of-words approach. It maps the keyword to its corresponding Ontology concept and calculates mapping score (m). This score (m) is used to make a concept document matrix and given as input to SVM classifier. The use of Ontology to create bag-of-concepts improved the classification of documents. Turney [2] discusses three approaches to document representation in natural language processing- term document vector (tdv), word context vector (wcv) as used in [9] and pair pattern vector (ppv). The tdv based methods find document similarity on the basis of frequency of terms in document. The wcv uses word context into consideration and accordingly the frequency is calculated. In ppv relationship between words is found along with the context using a text similarity score calculation method and using reasoning methods. In this paper, the domain Ontology is used by constructing concept map and calculating distance between two concepts.

Table 1. Survey of Subjective Evaluation techniques

| **Year** | **Author** | **Tool** | **Technique** | **Results** | **References** |
|---|---|---|---|---|---|
| Clustering technique | | | | | |
| 2003 | Landauer | Inteliigent Essay Assessor | Latent Semantic Processing | 59-88% | [3],[4]–[7] |
| 2008 | Kakkonen | Automatic essay Assessor | LSA, Probabilistic LSA, Latent Dirichlet Allocation | LSA better than rest | [8],[9], [10] |
| 2010 | Islam | | Generalized Latent Semantic Analysis | 86-96% | [11] |
| Classification techniques | | | | | |
| 2002 | Rudner | Betsy | Bays Theorem | 80% | [12] |
| 2008 | Li bin | | K-Nearest Neighbor | 76% | [13] |
| 2012 | Sukkarieh | C-rater | Maximum Entropy | 80% | [14]–[16] |
| Natural Language Processing based Techniques | | | | | |
| 1998 | Burstein | E-rater | Hybrid of features | 84-94% | [17] |
| 2001 | Callear | Automated Text Marker | Conceptual Dependency | None | [18] |
| 2005 | Perez | Atenea | BiLingual Evaluation Understudy, LSA | 50% | [19], [20] |
| Soft computing techniques | | | | | |
| 2009 | Wang | | Neural Network and LSA | Applied to information Retrieval only. | [21] |
| Semantic Techniques | | | | | |
| 2008-2012 | | | Ontology Based Methods | Applied to Information Retrieval Only | [1][2][22] |

## 3. Application Of Machine Learning Techniques To Subjective Evaluation Without Ontology

The machine learning techniques as used in implementing this work are discussed below. The input to all the techniques is keywords in the form of Model Answer and all Student Answers. The output is a similarity measure in the range of [0, 1], where a value of 0 indicates no similarity and 1 indicates high similarity. The input is pre-processed before applying techniques. The steps of pre-processing are: tokenization (find all words in all student answers), stop word removal (removing common words like a, the, as, an etc.), synonym search (for each word left after stop word removal, find its synonyms) and stemming (reduce the words to their stem).

Latent Semantic Analysis (LSA): The first step is to construct the Term-Document Frequency (tdf) matrix. To calculate tdf matrix, the complete diction of possible words is found by collecting all unique words in model answer and all student answers- called as masterTermVector. Then, we calculate tdf by counting the number of times each word in masterTermVector appears in each student's answer. Singular Value Decomposition is performed on the tdf matrix. It represents the individual words and all student answers as vectors. The model weight is found by adding the word vectors of all keywords in model answer. Vector Product of model weight and each student's answer vector is the similarity score.

Generalized Latent Semantic Analysis (GLSA): GLSA finds the diction by generating phrases (ngrams) from model answer and all student answers. The phrase length is 2, 3 and 4 neighboring words using sliding window for NGRAM1, NGRAM2 and NGRAM3 respectively. These phrases constitute the masterTermVector. The frequency of each phrase in each student's answer is calculated to generate tdf. The remaining steps are same as in LSA.

BiLingual Evaluation Understudy (BLEU): The technique calculates frequency of each word in each student's answer (sf) and keywords in model answer (kf). Ratio between sf and total number of words in each student's answer is calculated for each word. If sf value is larger than kf, then kf is used in this word average. The ratio is summed up for all the keywords in each student's answer to generate BLEU value.

Maximum Entropy (MAXENT): In MaxEnt, the input is training essays (multiple model answers) and all student answers. No pre-processing steps are performed. It uses a Perceptron to evaluate answers. The training data is used to study the word context – that is words that mostly follow and precede the word under consideration. The entropy is calculated for the current word to appear in a given context. This entropy is calculated for each word in model answer. The entropy for all word pairs in the model answers and corresponding target marks are given to Perceptron for training and then it is used for all student answers evaluation. Then it reads one student's answer at a time and finds if student's answer entails the standard answer concepts.

# 4. APPLICATION OF MACHINE LEARNING TECHNIQUES TO SUBJECTIVE EVALUATION WITH ONTOLOGY

## 4.1. DESIGN OF ONTOLOGY

The Domain Ontology of Computer Graphics is prepared using subject-predicate-object representation. The following components of Ontology are defined:

**1) Classes:** Sets, collections, concepts and types of objects. For example, the main class is Computer_graphics. Then Computer_graphics has computer_graphics applications, Computer_graphics_systems, types_of_media, etc. Then further classification of each is done as depicted in Figure-1.
**2) Individuals:** The classes have instances or objects as shown in Firgure-2. For example, image_processing class has individuals like color_coding, improve_picture_quality, improve_quality_of_shading, machine_perception and rearrange_picture_parts. The instances then further have properties.
**3) Attributes:** The classes and individuals have properties as shown in Figure-3. Some of the attributes identified are: isDefinedBy, uses, technique, purpose, standard, created_using, etc.
4) Relations among Classes: Apart from subclass relation, Ontology uses disjoint and equivalence relations.
**5) Process representation:** There are many processes and algorithms in graphics for example, working_of_CRT, working_of_plasma_panel, etc. These are represented as event class individuals as shown in Figure-4. These have properties like actor, target, output_of_event, input_to_event, part_of_event, predecessor, successor etc.

## 4.2. Usage Of Ontology

When the students' answers for the question are to be evaluated, the details related to the concept are fetched from the Ontology. The level of detail will depend on the type of question as shown in Table-2. When short questions are to be answered then directly related information is fetched. When longer questions are to be answered then more details are fetched. After the information is extracted from Ontology, a Multi-Hash map is created by collecting all the words corresponding to same concept. This Multi-Hash map is then used in evaluation. Apart from fetching the concept information from Ontology, the relation or similarity score between concepts is also fetched. If the concepts have a path between each other, then the length of such path is calculated. The combining of Ontology with the techniques mentioned in Section 3 is done using the design depicted in Figure 5. After performing preprocessing of words, the model answer and students' answers are given as input.

Table 2. Criteria to fetch information from Ontology

| Type of Question | Level of Detail fetched from Ontology |
|---|---|
| One-line questions | Properties of Concepts, instances and subclasses. |
| Short Length questions | All the concepts under the main concept. |
| Long Answer Questions, stating facts and phenomenon | All the concepts under the main concept along with equivalent and inverse classes. |
| Essay Length questions, reflective and open ended | All the concepts directly under the main concept and related concepts with inverse, equivalence, part-of, steps in phenomenon, etc. are fetched. All the nodes directly related to main concept and indirectly related are fetched in the form of Triples. |

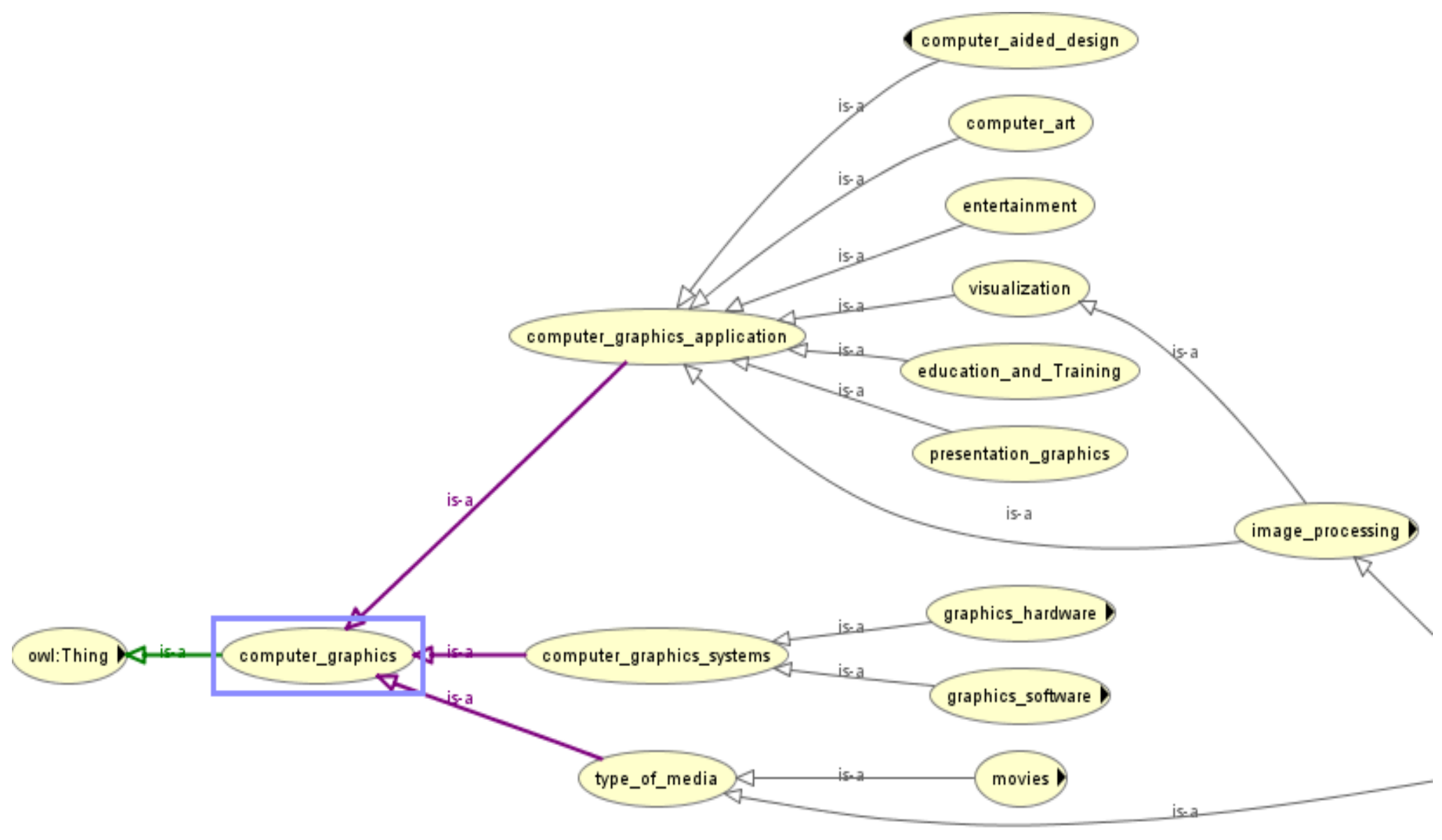

Figure 1. Part-of Hierarchy of classes defined in Computer Graphics Ontology

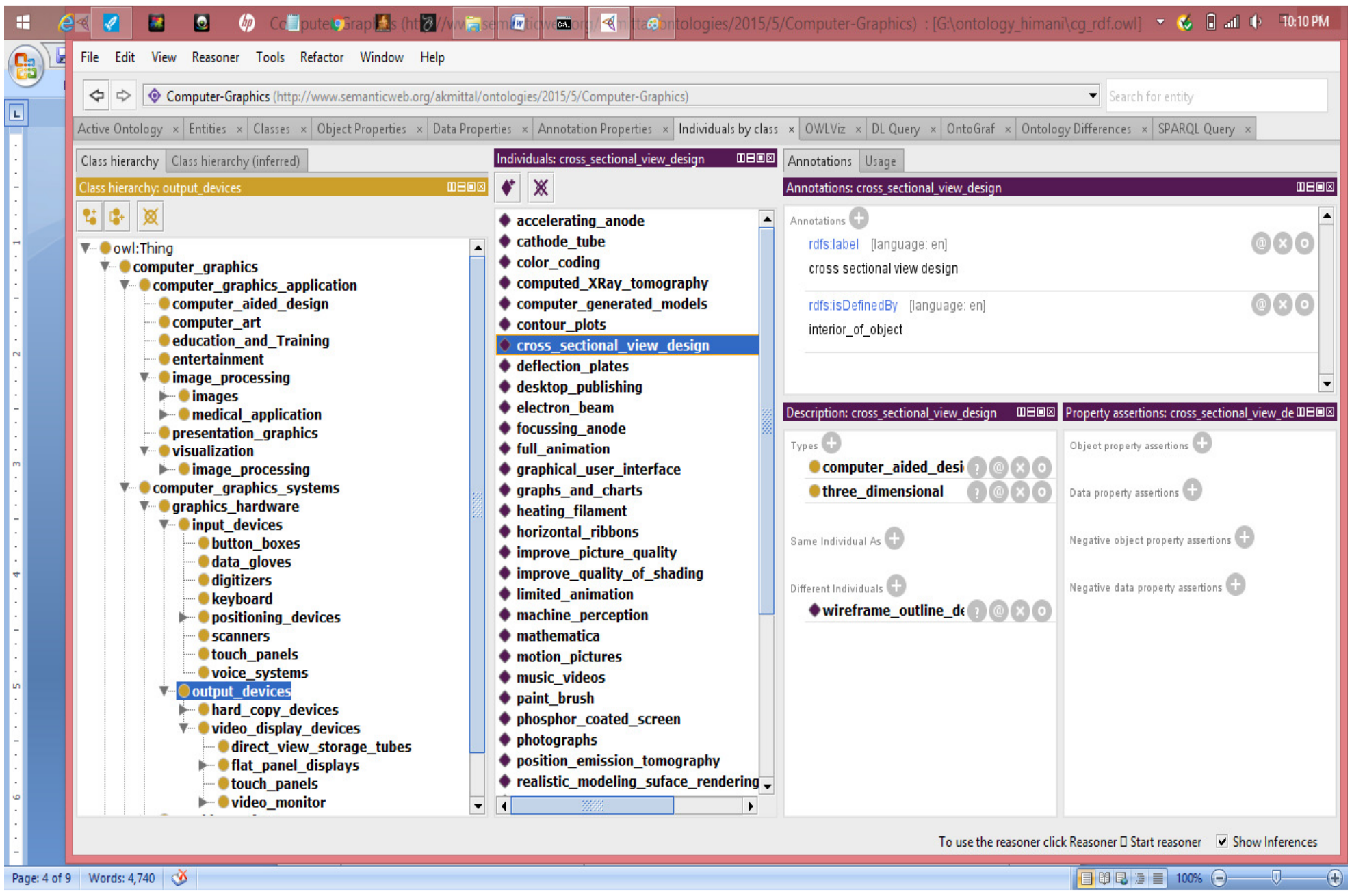

Figure 2. Individuals of Classes in Computer Graphics Ontology

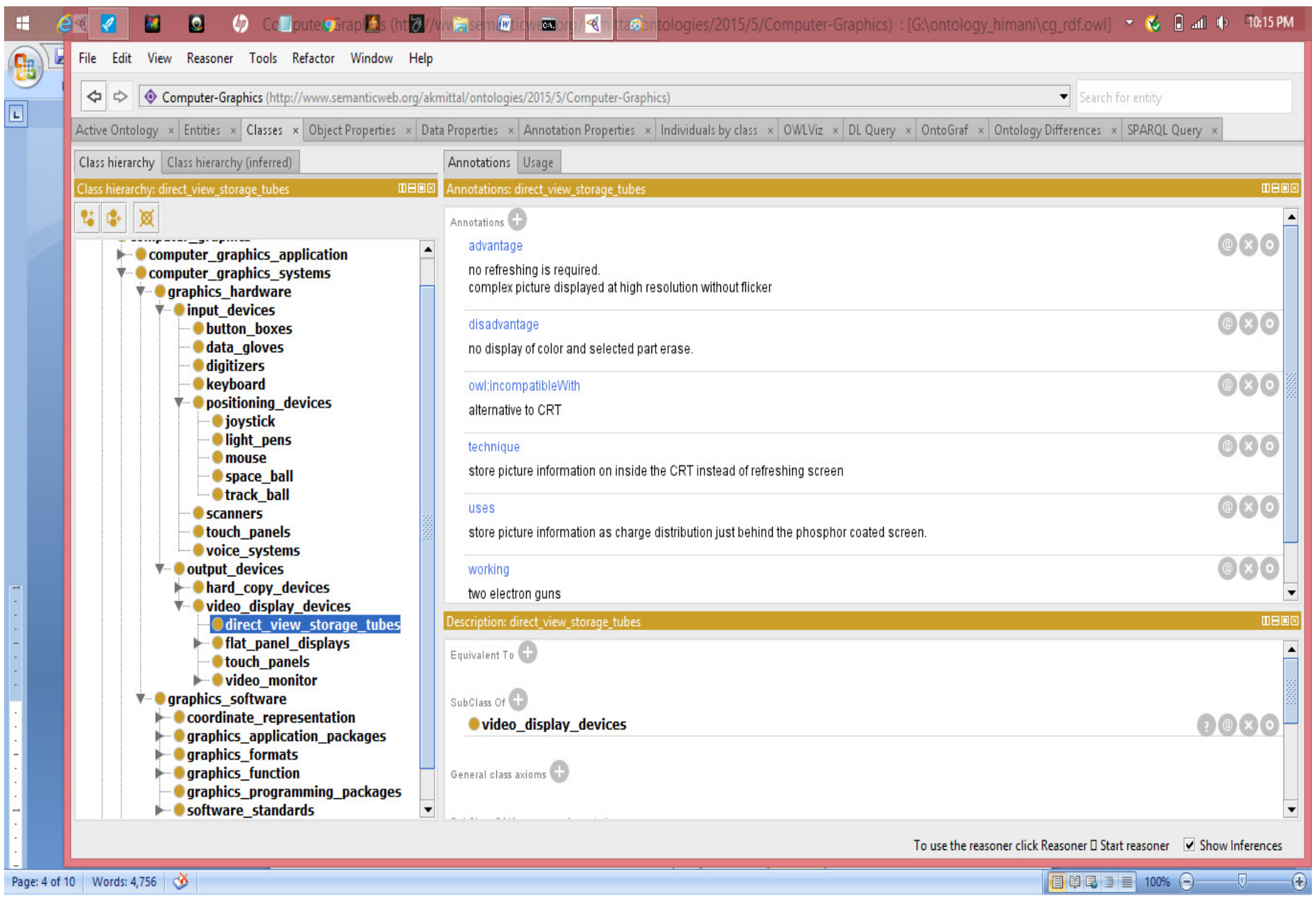

Figure 3. Properties of Classes in Computer Graphics Ontology

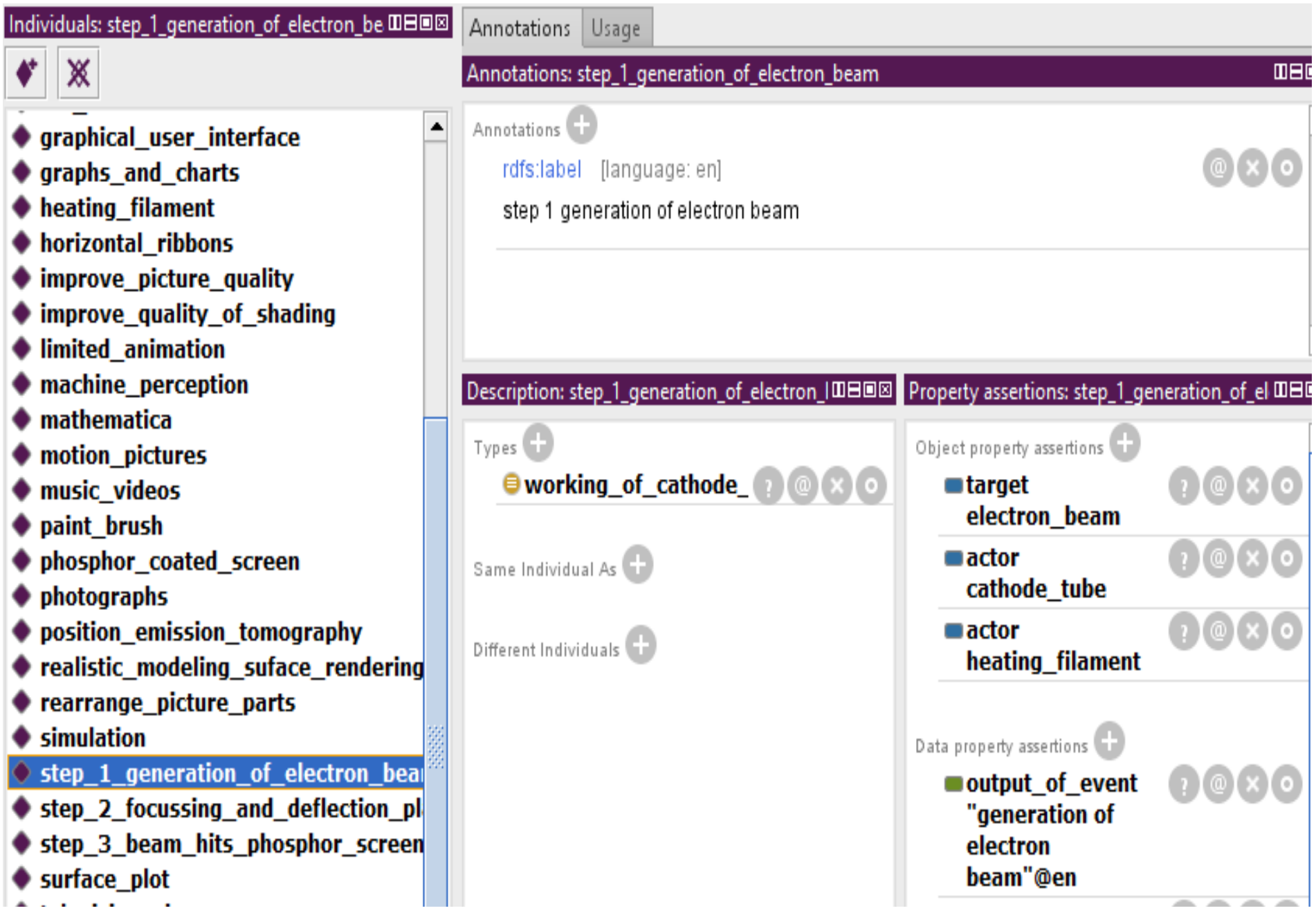

Figure 4. Event representation in Computer Graphics Ontology

The Ontology is extracted for the concept on which question is based. The sentences in model answer are then classified as belonging to these concepts. Then the new extended version of Ontology concept map is used for finding similarity between concept map and students' answers. The path length between various Ontology concepts is used to decide the relative importance of each concept in Ontology and weight to be associated with this concept.

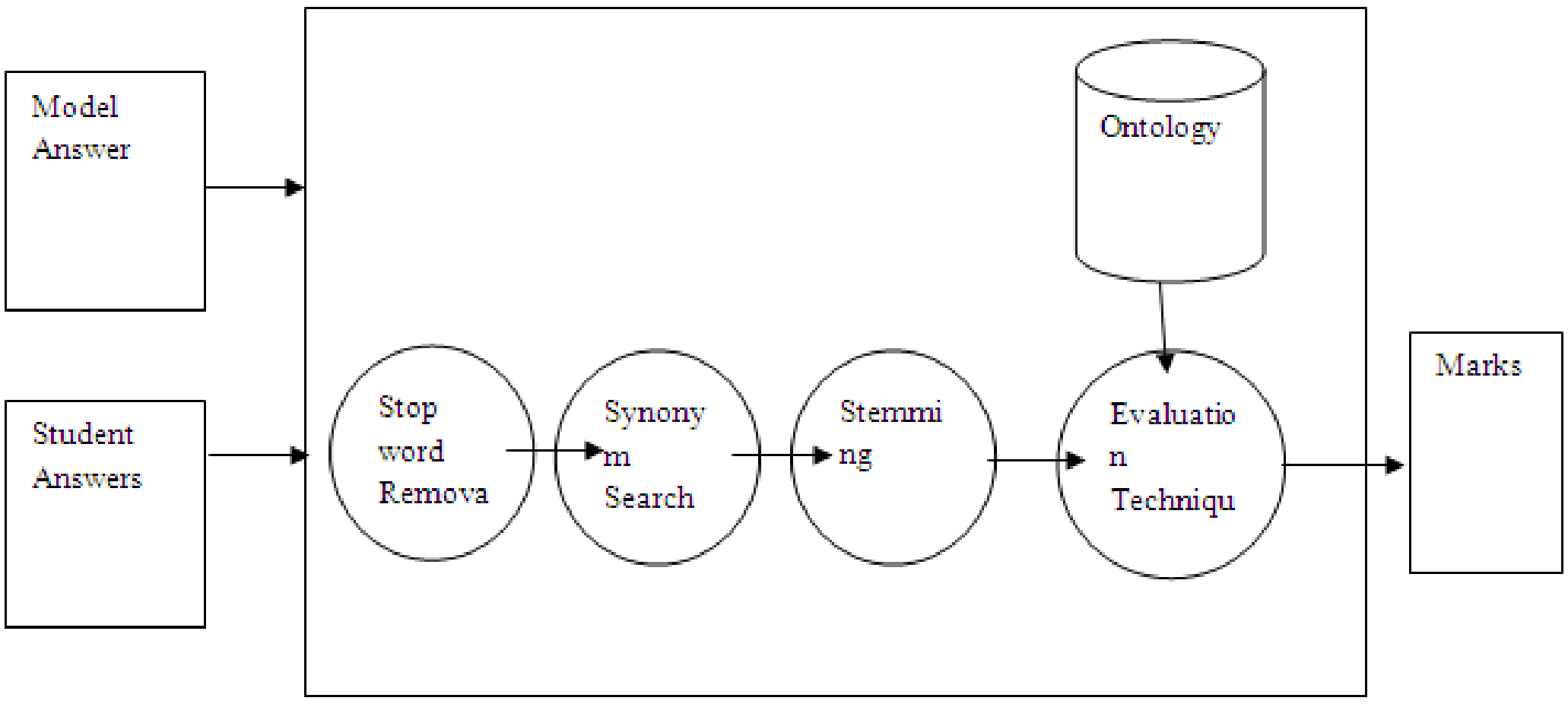

Figure 5. Design of evaluation methodology with ontology

## 4.3. Modification To Existing Techniques

The following steps are performed for each technique discussed in Section-3. The technique takes as input- multi-hash map of Ontology concepts, distance between concepts, model answer and students' answers. First the sentences in model answer are clustered using Ontology concepts and merged with the Ontology map using the machine learning technique in consideration. Then the updated multi-Hash map is used to find correlation between each concept and students' answer using same machine learning technique. The total number of concepts having positive correlation with students' answers (q) is multiplied by distance between the main concept and current concept. Then this value is divided by total number of concepts in multi-Hash Map to generate final score.

Along with the techniques mentioned in the above section, one more technique is applied with Ontology, the word-weight technique. In word-weight technique, after the words are fetched from Ontology, then words in model answer are combined with Ontology concepts and then weight of each keyword is calculated as – total occurrence of each word in students' answer divided by total number of words in students' answer.

## 5. IMPLEMENTATION

LSA technique is implemented in Java programming language and MatLab. MatLab is used for performing Singular Value Decomposition (SVD) for LSA. Open source libraries are used for invoking MatLab [24] from java code. Synonym Search is done using WordNet [25];[26]. GLSA is implemented by extending LSA package by incorporating n-grams. All the programming and extension tools used are same as in LSA. BLEU is implemented in java programming language. The Java based Maximum Entropy package is freely available at http://maxent.sourceforge.net. The features and working of this package are understood and it is used for evaluation. Word-weight technique is implemented in Java. The details of tools and techniques used for implementation is given in Table 3. Subject Specific Ontology was implemented for a subject of computer science - Computer Graphics. The format used for Ontology is RDF and tool used to construct Ontology is Protégé 5.0 beta. In order to incorporate Ontology in the techniques, information in Ontology is extracted using Sparql query 1.1 and Apache Jena Library 2.13.0.

Table 3. Tools and Techniques used for subjective evaluation

| Phase | Tool | Library | Technology/ Technique | Purpose |
|---|---|---|---|---|
| **Prototype** | Java Development Kit 1.7 | | | It is used for development of Whole application |
| **Subjective Evaluation** | | Stemmer | | Porter's Algorithm implemented by originator of the algorithm |
| | MatLab 2013 | Matlab Control Library | Matrix Operations like singular valued decomposition | Used for LSA matrix calculations. |
| | WordNet 2.1 | JWI 2.2.3 | Semantic Networks | Used for finding word Synonyms |
| | | Guava library | Multi Hash Lists | Used for counting frequency of words |
| | Protégé 5.0 beta | Apache Jena library | RDF format of Ontology | Subject Specific Ontology Development |
| | | MaxEnt Library | Maximum Entropy | Downloaded package for performing evaluation |

## 6. TESTING AND RESULTS

There is no standard database to test subjective answer evaluation. Therefore, the database was created over a period of time by conducting class tests. All the techniques (as implemented above) for answer evaluation have been tested using this common database. The database consists of ten different questions with about sixty answers for each question from the domain of Computer Graphics. The marks are generated for each student's answer using the techniques with and without Ontology. Table-4 contains the correlations between human assigned score and machine generated score using each technique.

Table 4. Correlations of performance of different techniques with Human

| | BLEU | MAXENT | LSA | NGRAM1 | NGRAM2 | OWW | OBLEU | OMAX | OLSA | ONG1 | ONG2 |
|---|---|---|---|---|---|---|---|---|---|---|---|
| Max | 0.90 | 0.54 | 0.91 | 0.90 | 0.88 | 0.96 | 0.96 | 0.93 | 0.91 | 0.95 | 0.93 |
| Min | 0.71 | 0.29 | 0.69 | 0.34 | 0.34 | 0.29 | 0.71 | 0.66 | 0.61 | 0.34 | 0.34 |

## 7. ANALYSIS OF RESULTS

In this section, the results of each technique are analyzed. It is clear from the Table-4 that BLEU and LSA give the consistent performance with and without Ontology. With Ontology, Maximum Entropy shows a lot of improvement. LSA and GLSA, both overrate if keywords or phrases are repeated many times. BLEU technique is a word average technique and cannot capture the relationship between words. MaxEnt cannot identify the discrepant essays. BLEU identifies discrepant essays. Comparing NGRAM-1,2,3 with LSA, as they are theoretic improvements over LSA, do not give better results than LSA. Actually, they are giving a minimum performance. The use of n-gram size 1,2,3 gives almost same results. So, increasing the size of n-gram does not give better performance. With use of Ontology, all the techniques OBLEU, OLSA, ONG<1, 2, 3> and OMAX perform as good as original techniques. The improvement is not in the overall correlation but the individual marks assigned to each answer. The techniques discussed and implemented in this paper show a high correlation (upto 90 percent) with Human Performance. This is because human evaluation is by and large influenced by answer length, keyword presence and context of keywords. Without Ontology, only two levels of human mind modeling can be covered. Use of Ontology, on the other hand, looks not just for keywords but the keywords appearing in right context and thus models human mind more accurately.

In Table-5, all the techniques are compared on the basis of a number of parameters as selected from the working and definition of these techniques. A technique should have properties of Semantic study, negative-positive role, syntactic importance and discrepant essay identification; and should not have bag-of-words property. OMAX has best performance in semantic study (builds concept model for training of neural network), negative-positive role (takes care of word order and identifies yes and no roles) and syntactic importance (parses the sentences and finds importance of each word).

## 8. CONCLUSION AND SCOPE FOR FUTURE WORK

The techniques discussed and implemented in this paper show a high correlation (upto 90 percent) with Human Performance. This is because human evaluation is by and large influenced by answer length. keyword presence and context of keywords. Use of Ontology, checks for not only keywords but for the keywords appearing in right context. This aspect is lacking in techniques used without Ontology. The use of Ontology checks for presence of keywords, synonyms, right word context and coverage of all concepts. It is concluded that using ML techniques with Ontology gives satisfactory results due to holistic evaluation. This work can be enhanced by using extended Ontology including the concepts of all computer science subjects. It can be further improved to provide students feedback about the missing concepts in their answers.

Table 5. Performance comparison of techniques used

| Criteria | Absolute Yes | YES | Intermediate | NO |
|---|---|---|---|---|
| **Semantic study** | OMAX (builds concept model) | LSA,OLSA, GLSA, OGLSA, MAX (build model of answers on basis of keywords) | | BLEU, OBLEU and OWW (since they are word average techniques) |
| **Negative-Positive role** | MAX, OMAX (take word order into consideration and negative role specifically attached to the word itself) | GLSA and OGLSA (take word order in consideration) | | BLEU, OBLEU, OWW, LSA,OLSA (no importance to word order and negative role words are treated independent of original words) |
| **Syntactic Importance** | OMAX, MAX (parsing is done) | | GLSA, OGLSA | BLEU, OBLEU, OWW, LSA, OLSA |
| **Bag of words** | LSA, OLSA, BLEU, OBLEU, OWW | | | MAX, OMAX, GLSA, OGLSA |
| **Time taking** | OGLSA, GLSA (takes 15 minutes to execute) | | LSA, OLSA | BLEU, OBLEU, OWW, MAX, OMAX |
| **Discrepant Essay Identification** | | BLEU, OBLEU | | OWW, MAX, GLSA, OGLSA, LSA, OLSA |